\documentclass{article}

\usepackage[final, nonatbib]{neurips2021}

\usepackage[utf8]{inputenc} 
\usepackage[T1]{fontenc}    
\usepackage{url}            
\usepackage{booktabs}       
\usepackage{amsfonts}       
\usepackage{nicefrac}       
\usepackage{microtype}      
\usepackage{graphicx}
\usepackage{amsmath,mathtools}
\usepackage[table,xcdraw]{xcolor}
\usepackage{tabularx}
\usepackage{wrapfig}

\renewcommand{\footnotesize}{\fontsize{9.2pt}{11pt}\selectfont}
\usepackage[colorlinks=true, linkcolor=red]{hyperref}

\title{Self-Supervised Vision Transformers Learn 
\\ Visual Concepts in Histopathology}
\author{%
  Richard J. Chen$^{*}$ \\
  Department of Biomedical Informatics \\
  Department of Pathology \\
  Harvard Medical School \\
  \And
  Rahul G. Krishnan$^{*}$ \\
  Department of Computer Science \\
  Department of Laboratory Medicine and Pathobiology \\
  University of Toronto \\
}
    
\begin{document}

\maketitle

\begin{abstract}
\let\thefootnote\relax\footnote{$^{*}$ Part of this work performed while at Microsoft Research New England.}
Tissue phenotyping is a fundamental task in learning objective characterizations of histopathologic biomarkers within the tumor-immune microenvironment in cancer pathology. However, whole-slide imaging (WSI) is a complex computer vision in which: 1) WSIs have enormous image resolutions with precludes large-scale pixel-level efforts in data curation, and 2) diversity of morphological phenotypes results in inter- and intra-observer variability in tissue labeling. To address these limitations, current efforts have proposed using pretrained image encoders (transfer learning from ImageNet, self-supervised pretraining) in extracting morphological features from pathology, but have not been extensively validated. In this work, we conduct a search for good representations in pathology by training a variety of self-supervised models with validation on a variety of weakly-supervised and patch-level tasks. Our key finding is in discovering that Vision Transformers using DINO-based knowledge distillation are able to learn data-efficient and interpretable features in histology images wherein the different attention heads learn distinct morphological phenotypes. We make evaluation code and pretrained weights publicly-available at: \href{https://github.com/Richarizardd/Self-Supervised-ViT-Path}{https://github.com/Richarizardd/Self-Supervised-ViT-Path}.
\end{abstract}

\section{Introduction}
Tissue phenotyping is a fundamental problem in computational pathology that aims at learning objective characterizations of histopathologic biomarkers within the tumor-immune microenvironment for cancer diagnosis, prognosis, and response-to-treatment \cite{ludwig2005biomarkers, kather2016multi, javed2020cellular}. In cancer pathology, the current clinical paradigm is the manual and subjective interpretation of histology whole-slide images (WSI) in determining clinical endpoints such as cancer subtype, grade, and stage~\cite{amin2017eighth}. Though used as the gold standard in taxonomic classifications and staging systems for many cancer types, such subjective interpretation has been demonstrated to suffer from large inter- and intra-observer variability~\cite{nicholson2001reproducibility, rabe2019interobserver}. For instance, the substantial variability in discerning well-to-poorly differentiated glands in the Gleason grading system for prostate adenocarcinoma, or the evaluation of nuclear size, shape, and nucleolar prominence in the Furhman grading system for renal cell carcinoma~\cite{carlson1998accuracy, novara2007grading}. Though current efforts using deep learning have progressed the field towards using Convolutional Neural Networks (CNNs) with pixel-level annotations, the large image resolutions of WSIs add cost to annotation workflows and precludes the development of tissue phenotyping algorithms at scale in addressing diverse tasks.

To address the limitations of using pixel-level annotations for tissue phenotyping, recent progress in set-based deep learning have adopted frameworks such as multiple instance learning (MIL) for weak-supervision of WSIs using only slide- or patient-level labels. Such frameworks follow two steps: 1) instance-level patch embeddings are extracted from non-overlapping tissue patches in the WSI, and then 2) global pooling is applied over the bag to construct a WSI-level embedding~\cite{campanella2019clinical}. This work studies the first of those two steps in building predictive models from histopathological images. In the status quo, there is a lack of diverse and well-curated pathology datasets that would enable generalization across diverse tissue and organ types in cancer pathology. Consequently, a common choice made for the instance-level feature extractor is a ResNet-50 encoder pretrained on the ImageNet dataset. Although a reasonable starting point, this model which may not capture all the domain-specific features relevant in downstream tasks for cancer pathology~\cite{lu2020data} due to the differences between histopathological images and natural images.

\begin{figure}
   \includegraphics[width=14cm]{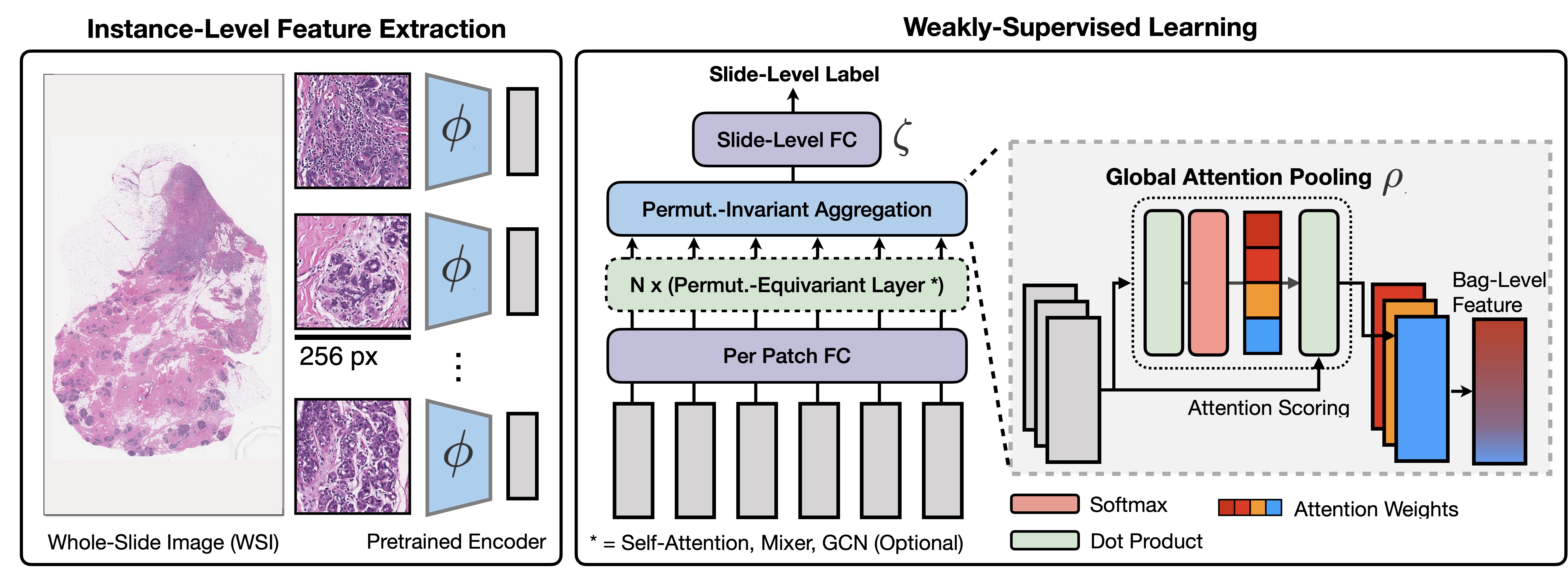}
   \caption{Conventional weak-supervision in WSIs using multiple instance learning and set-based deep learning. Feature embeddings are extracted using an encoding function $\phi$ (pretrained ResNet50 on ImageNet after the 3rd residual block), with only $\rho, \psi$ trained in feature aggregation ($\rho$ typically implemented as global attention pooling from Ilse \textit{et al.}. Permutation-equivariant functions (\textit{e.g.} - MLP-Mixer, Transformer attention, graph convolutions) can be optionally stacked before pooling)~\cite{tolstikhin2021mlp, vaswani2017attention,chen2021multimodal}.}
\end{figure}

More recently, deep learning has witnessed a surge in self-supervised learning ~\cite{oord2018representation, donahue2019large, chen2020simple, he2020momentum, grill2020bootstrap, chen2021exploring, caron2021emerging} as a paradigm for learning feature representations without the use of labels. Self-supervised
learning leverages the insight that one can learn useful representations of high-dimensional data through the use of auxiliary tasks such as identifying that the representation of an image (under a information preserving transformation) should not change very much. While self-supervised learning has been proposed as a replacement for ResNet-50 encoders pretrained on ImageNet \cite{deng2009imagenet, he2016deep} in pathology tasks, we note two limitations of existing work. The first is a lack of comprehensive benchmarks that evaluate self-supervised models on diverse patch-level and weakly-supervised tasks, and the second is a lack of introspection and post-hoc assessment of the learned self-supervised representations in understanding what morphological features are learned.

The overarching objective of this work is to learn good representations for WSIs and understand what aspects of histopathological images are captured by such representations. We highlight the following contributions of our work: 
\begin{enumerate}
\item We train several state-of-the-art self-supervised learning models and apply them to a diverse variety of patch-level and weakly-supervised tissue phenotyping tasks in computational pathology. In doing so, we study how the inductive biases of these techniques affect the quality of representations in downstream tasks. Our benchmark evaluates two different self-supervised models (SimCLR, DINO) trained on the breast invasive carcinoma (BRCA) cohort in The Cancer Genome Atlas (TCGA). 
\item We find from analysis of our baselines that ResNet-50 encoders pretrained on ImageNet work better than previously reported, achieving comparable performance to a several self-supervised pretraining methods.
\item Next, we show that self-supervised learning via knowledge distillation in DINO \cite{caron2021emerging} achieves the highest performance across a majority of tasks. We believe that this improvement comes from potential local-global correspondences captured in DINO data augmentation, which may model part-whole hierarchies inherent in histopathology imaging data (\textit{e.g.} - local image crops describing cells in the global tissue patch).
\item Finally, we find that DINO learn visual concepts of histopathology tissue in localizing stroma tissue, cell location, and regions of fat / air pockets with different attention heads, suggesting that ViTs are able to capture important inductive biases about histopathology tissue.
\end{enumerate} 

\section{Related Work}

\textbf{Unsupervised Learning for Tissue Phenotyping.} Recent work in unsupervised deep learning has focused on devising proxy objectives to train deep networks in the absence of target labels, demonstrating impressive performance on image representation learning on both natural images and histology tissue patches. Oord \textit{et al.} proposed one of the first methods in self-supervised learning using contrastive losses (called Contrastive Predictive Coding, or CPC), which aims to maximize the mutual information between $64 \times 64$ image patches within a $256 \times 256$ image (positive target), and minimize the mutual information between image patches of different images in the mini-batch (negative target)~\cite{oord2018representation}. Similarly, Chen \textit{et al.} proposed a contrastive framework that maximizes the cosine similarity between two data-augmented views of the same image, and minimizes similarity between different images within the mini-batch~\cite{chen2020simple}. In pathology applications, CPC, SimCLR, and other similar self-supervised methods have been used for cell- and tissue-level phenotyping, with downstream applications in cancer diagnosis, cancer prognosis, and gene mutation prediction~\cite{chen2020pathomic, dehaene2020self, zhao2020predicting, ciga2020self, li2021dual, koohbanani2021self, ciga2021resource, saillard2021self, claudio2021adversarial, ciga2020self, srinidhi2022self}. However, an important limitation is that though achieving strong performance on standard benchmarks such as ImageNet and STL-10, these methods depend on strong assumptions about the class balance of the dataset, as sampled minibatches would typically sample images of different classes in these datasets. Such assumptions may not hold in WSIs, as depending on the cancer type, there may exist large class imbalances of morphological phenotypes as sampled histology patches may comprise of only one class and have similar statistical image properties (e.g. - only benign tissue)~\cite{chuang2020debiased}. In addition to contrastive learning, self-supervised learning via student-teacher knowledge distillation frameworks have recently emerged in achieving state-of-the-art results on ImageNet Top-1 and Top-5 benchmarks~\cite{grill2020bootstrap, caron2021emerging, wang2021transpath}. An important distinction in comparison to contrastive learning is that these methods can be trained without the need of negative targets, however, large-scale efforts in benchmarking knowledge distillation self-supervision in computational pathology have had limited attention.

\textbf{Weakly-Supervised Learning in WSIs.} Remarkable progress have been made in weakly-supervised learning in WSIs using set-based network architectures. In general set-based deep learning, Edwards and Storkey and Zaheer \textit{et al.} proposed the first network architectures operating on set-based data structures~\cite{edwards2016towards, zaheer2017deep}. In pathology, Ilse \textit{et al.} extended set-based network architectures as an approach for multiple instance learning in histology region-of-interests, with Campanella \textit{et al.} later extending end-to-end weak-supervision on gigapixel WSIs~\cite{ilse2018attention, campanella2019clinical}. Lu \textit{et al.} demonstrated that by using a pretrained ResNet-50 encoder on ImageNet at the instance-level feature extraction step, only a global pooling operator needs to be trained in reaching state-of-the-art performance on many diagnostic subtyping tasks~\cite{lu2020data}. Following the work of Lu \textit{et al.}, there have been many variations of MIL that have adapted unsupervised learning techniques such as VAE-GAN, SimCLR, and MOCO as instance-level feature extraction~\cite{zhao2020predicting, chen2020pathomic, li2021dual, saillard2021self}. However, we note that there is a lack of comprehensive benchmarks that evaluate self-supervised models on diverse patch- and slide-level tasks.

\section{Background}

\textbf{Notation:} For a given WSI, let $\mathbf{x}$ refer to a $256 \times 256$ tissue-containing image patch at $20\times$ magnification, with the bag of the $M$ tissue-containing patches in the WSI as $\mathbf{X} = \{\mathbf{x}_{1}, \dots, \mathbf{x}_{M}\} \in \mathbb{R} ^ {M \times 3 \times 256 \times 256}$. Let $Y$ denote the slide-level label corresponding to $\mathbf{X}$, and $y$ refer to the patch-level label corresponding to $\mathbf{x}$. 
We use $\phi: \mathbb{R}^{3 \times 256 \times 256} \rightarrow \mathbb{R}^{1 \times d}$ to denote an encoder that extracts $d$-dim features from $\mathbf{x}$. 
After applying $\phi$ per-patch we obtain a bag of extracted features $\mathbf{H} = \{\mathbf{h}_{1}, \dots, \mathbf{h}_{M}\} \in \mathbb{R} ^ {M \times d}$.

\textbf{Slide-Level Classification.} Multiple Instance Learning (MIL) is a framework that operates on bags of embedding instances, in which label information is provided at the bag-level but not the instance-level. The primary goal is to solve the bag classification task $P(Y|\mathbf{X})$, with the additional challenge of discovering key instances $\mathbf{x}$ that would trigger the bag label. As a set-based network architecture, we train a permutation-invariant function $\mathcal{F}$ that has the general form:
\begin{equation}
\mathcal{F}\left(X\right)= \zeta \big(\rho \left( \{ \phi \left(\mathbf{x}_i\right): \mathbf{x}_i \in \mathbf{X} \} \right) \big)
\end{equation}
which learns: 1) the encoder $\phi$ applied instance-level extracting $\mathbf{h}$ from $\mathbf{x}$ in the bag, 2) the symmetric, permutation-invariant aggregation function $\rho: \mathbb{R}^{m \times d} \rightarrow \mathbb{R}^{1 \times d}$ that pools $\mathbf{h}$ in the bag as the bag-level feature $\mathbf{H}$, and 3) a bag-level classifier $\zeta: \mathbb{R}^{d} \rightarrow \mathbb{R}^{\text{\# class}}$ that further processes the bag-level features for downstream classification tasks. In current frameworks for MIL in WSIs, training $\mathcal{F}$ is usually not performed end-to-end due to computational complexity of gigapixel image resolutions. In Lu \textit{et al.}, $\phi$ is implemented as a pretrained ResNet-50 model on ImageNet truncated after the 3rd residual block (denoted as ResNet-50-B3$_{\text{IN}}$) extracting 1024-dim features (\textbf{Figure 1})~\cite{lu2020data}. As a result, only $\rho, \zeta$ are trained in MIL, typically parameterized using the global attention pooling function in Ilse \textit{et al.}~\cite{ilse2018attention}.

\textbf{Patch-Level Classification:} Using pixel-level annotations $y$, extracted feature embeddings using $\phi$ can then be finetuned for simple patch classification $P(y|\mathbf{x})$, in which $\phi$ is a previously pretrained ResNet-50 or ViT model. We note that though tissue phenotyping can be formulated as a segmentation problem, due to the gigapixel image resolutions of WSIs, patches contained within segmentation masks would typically belong to one class. For some tasks such as cell type classification in which patches can have an admixture of diverse phenotypes, we use the majority phenotype label as the patch label~\cite{amgad2019structured}.

\section{Self-Supervised learning as Instance-Level Pretraining}

\begin{figure}[ht!]
   \begin{center}
   \includegraphics[width=14cm]{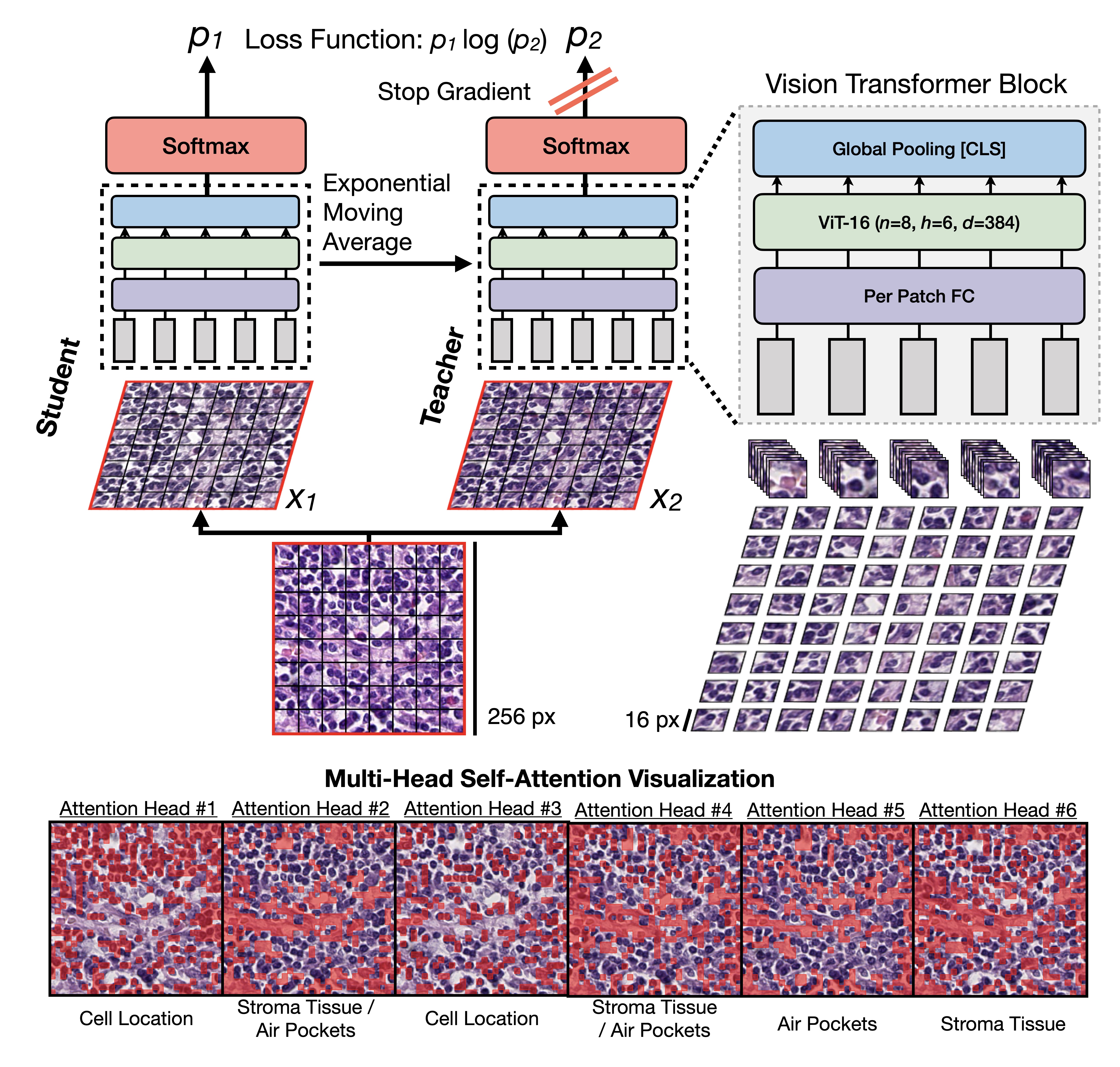}
   \end{center}
   \vspace{-3mm}
   \caption{Self-supervision using knowledge distillation in DINO to pretrain $\phi$ on histology image patches~\cite{caron2021emerging}. A student network $\phi_s$ is trained to match the probability distribution of a Siamese teacher network $\phi_t$ using a cross-entropy loss, with $\phi$ parameterized using a Vision Transformer (ViT) model, and local and global crops applied as data augmentation. Interpretability of multi-head attention weights reveal that DINO learns distinct morphological phenotypes.}
\end{figure}

While using ResNet-50-B3$_{\text{IN}}$ as the instance-level feature extraction $\phi$ has been successful in many weakly-supervised tasks, the extracted feature embeddings from an out-of-domain classifier may not generalize well across phenotyping tasks in cancer pathology. In this work, we study several self-supervised methods to pretrain $\phi$ for downstream tasks, and explore the quality of representations of SimCLR and DINO. Moreover, we note several unique properties about DINO pretraining, which we note below in the application of image pretraining for pathology data.

\textbf{SimCLR and DINO Overview:} Self-supervised approaches work by taking an image $\textbf{x}$, an augmented image $\tilde{\textbf{x}}$ (where $\mathcal{T}:\textbf{x}\to\tilde{\textbf{x}}$ is an image transformation that preserves information content e.g. a slight rotation) and comparing the representations $\phi(\textbf{x}),\phi(\tilde{\textbf{x}})$ in different ways. SimCLR learns the representation $\phi$ by minimizing
$\dfrac{\exp(\text{sim}[\phi(\textbf{x}_i),\phi(\tilde{\textbf{x}_i})])}{\sum_{k} \exp(\text{sim}[\phi(\tilde{\textbf{x}_i}),\phi(\textbf{x}_k)])}$, where $\textbf{x}_k=\{\textbf{x}_j\neq \textbf{x}_i;\textbf{x}_j\in \textbf{x}_{\text{minibatch}}\}$ and \emph{sim} is a similarity function (e.g. the dot product between two vectors). For images in $\textbf{x}_{\text{mini-batch}}$, the goal is to learn a representation that is robust to transformations that $\mathcal{T}$ that preserve the information content within the image, via maximizing similarity between data-augmented views of the same image (positive target) and minimizing similarity between views from different images in the mini-batch. As mentioned, however, this assumption may not hold for sampled mini-batches in pathology data which may have large class imbalance of represented morphological phenotypes. In DINO, rather than using a contrastive loss, student-teacher knowledge distillation is used, in which a student function $\phi_s$ is trained to match the probability distribution of a Siamese teacher network $\phi_t$ using the cross-entropy loss $-p_s(\mathbf{x}) \operatorname{log} p_t(\mathbf{x})$ with momentum encoding, with $p_s, p_t$ denoting the outputs of $\phi_s(\mathbf{x}), \phi_t(\mathbf{x})$ respectively for image $\mathbf{x}$ (\textbf{Figure 2})~\cite{caron2021emerging}. A desirable property of DINO in the application of self-supervised learning to pathology data is that negative targets are not needed to steer representation learning.

Different methods use different choices for $\mathcal{T}$. In SimCLR, color jittering, random crops and resizes, and horizontal flips used to augment the image while still preserving semantic information. DINO additionally constructs a set of 8 local views ($96 \times 96$ crops, passed through $\phi_s$) and 2 global views ($224 \times 224$ crops, passed through $\phi_t$) to encourage local-to-global correspondences between the student and teacher. An intriguing property that makes this data augmentation particularly well suited for histology data is the inherent part-whole hierarchy of cells in a tissue patch~\cite{hinton1988representing, chang2013classification, saltz2018spatial}. In comparison to natural images in which $96 \times 96$ crops may capture only colors and textures without any semantic information, at $20\times$, local $96 \times 96$ image patches encapsulate single cells and their surrounding extracellular matrices, which has shared mutual information with the broader spatial organization of cells in the global views.

\begin{figure}
   \includegraphics[width=14cm]{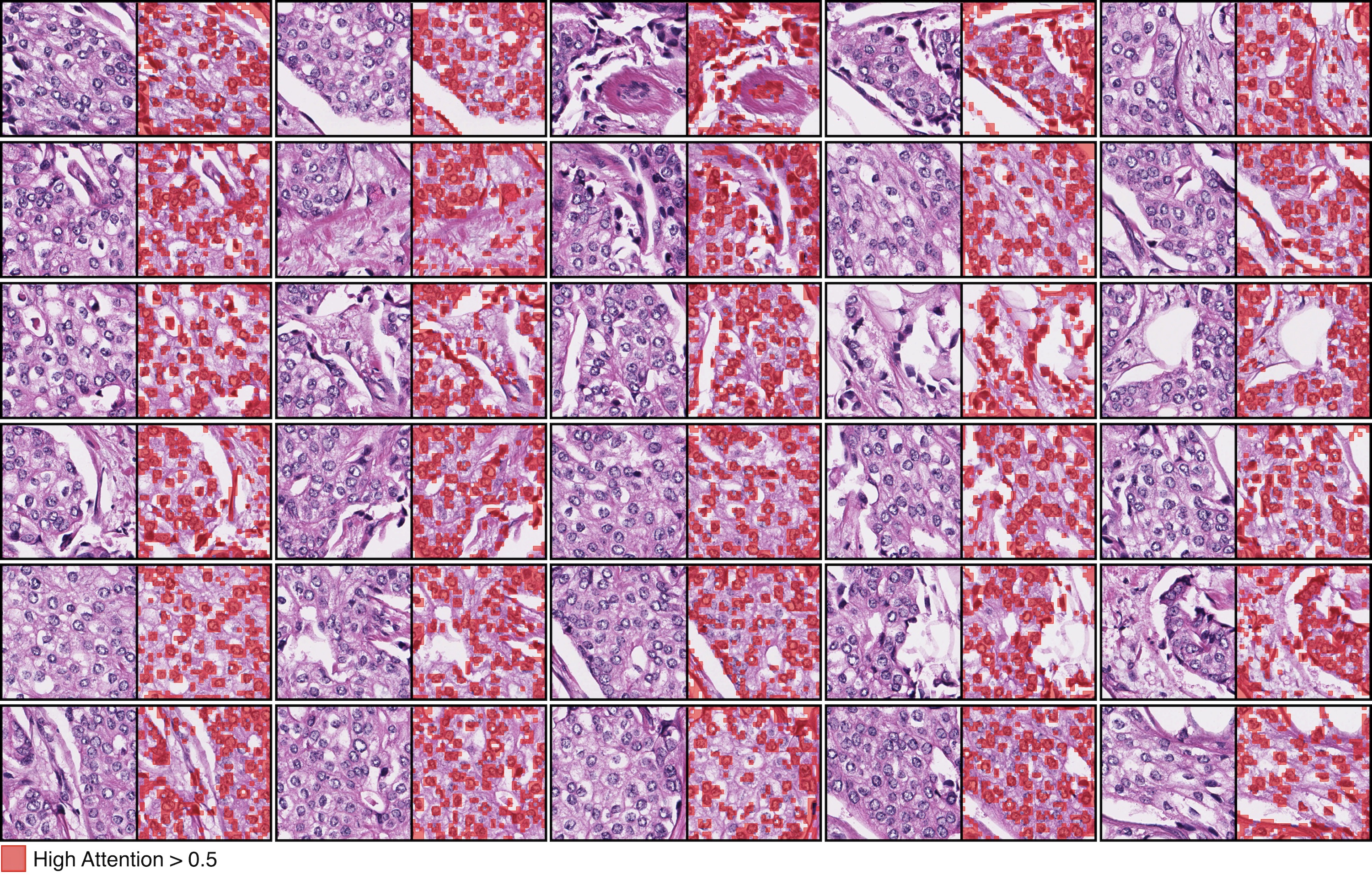}
   \vspace{-3mm}
   \caption{Self-Supervised ViTs localize cells as high-attention visual tokens. We visualize the attention weights for the first attention head, which we empirically found to be robust at localizing cells across a variety of histology image patches. Overlayed in "red" are high-attention $16 \times 16$ visual tokens with attention weights greater than 0.5.}
\end{figure} 

\textbf{Parameterizing $\phi$:} SimCLR uses a ResNet-50 encoder for $\phi$. In contrast, DINO uses a Vision Transformer (ViT) as the image encoder. As input, $256 \times 256$ image patches are further tessellated into non-overlapping $16 \times 16$ patches, with multi-head self-attention layers from Vaswani \textit{et al.} used as permutation-equivariant feature aggregation functions to learn interactions between the smaller patches. This choice allows the model to learn how cells are organized in the tissue patch~\cite{vaswani2017attention}. 


\section{Experiments and Results}

\subsection{Training $\phi$:}

To create the training datasets for image pretraining, we used the Tissue Image Analysis (TIA) toolbox to tessellate each WSI into non-overlapping $256 \times 256$ tissue-containing patches at $20\times$ magnification. In total, 2055742 image patches were curated from 1038 WSIs from the TCGA-BRCA cohort. All methods were developed on the same training datasets (of each organ type) with standard learning and data augmentation parameters of their respective source papers, and evaluated at 100 epochs. 

\subsection{Fine-tuning and evaluating $\phi$:}

We study a diverse array of tasks on which to assess and compare the different strategies for obtaining patch level embedding $\phi$. 

\textbf{Weakly-Supervised Cancer Subtyping:} As weakly-supervised evaluation, we used the TCGA-BRCA cohort for Invasive Ductal Carcinoma (IDC) versus Invasive Lobular Carcinoma (ILC) subtyping. Using the CLAM backbone as our set-based network architecture for feature aggregation, we trained CLAM with the following extracted embeddings: 1) ResNet-50-B3, 2) SimCLR, 3) DINO~\cite{lu2020data}. In addition, we also train all models with different percentage folds of the training dataset ($100\%/75\%/50\%25\%)$ as data efficiency experiments. For all models in weakly-supervised evaluation, we report the macro-averaged test AUC performance from 10-fold cross-validation.

\textbf{Patch-Level Tissue Phenotyping:} As tissue phenotyping tasks, we used the following datasets with patch-level annotations:

\begin{itemize}
    \item \textbf{CRC-100K:} CRC-100K is a dataset of 100,000 histological images of human colorectal cancer and healthy tissue, extracted as $224 \times 224$ patches at $20\times$ magnification, and is annotated with the following non-overlapping tissue classes: adipose (Adi), background (Back), debris (Deb), lymphocytes (Lym), mucus (Muc), smooth muscle (Mus), normal colon mucosa (Norm), cancer-associated stroma (Str), colorectal adenocarcinoma epithelium (Tum)~\cite{kather2016multi}. We experiment on CRC-100K with and without Macenko stain normalization (SN). As evaluation, we report one-versus-all AUC performance of each class as well as multiclass AUC performance.
    \item \textbf{BreastPathQ:} BreastPathQ is a challenge dataset from the TCGA-BRCA cohort that measures tumor cellularity, which measures the fractional occupancy of tumor cell presence in the image patch~\cite{petrick2021spie}. We evaluate on the public train/validation split of the challenge, which provides 2579 and 187 patches respectively at $20\times$, and report mean-squared error (MSE) and Kendall-Tau concordance.
\end{itemize}

\begin{table*}
\footnotesize
\begin{center}
\begin{tabular}{ll|cccc}
\toprule
{} & {} & \multicolumn{4}{c}{\underline{BRCA Subtyping}} \\
Method &       Arch &  100\% Training &              75\% Training &              50\% Training &              25\% Training \\
\midrule
IN Transfer & ResNet-50 & 0.884 $\pm$ 0.059 &  0.850 $\pm$ 0.069 &  0.835 $\pm$ 0.087 &  0.756 $\pm$ 0.081 \\
SimCLR &  ResNet-50 & 0.879 $\pm$ 0.069 &  \textbf{0.859 $\pm$ 0.079} &  0.820 $\pm$ 0.102 &  0.774 $\pm$ 0.094 \\
DINO & ViT &  \textbf{0.886 $\pm$ 0.059} &  0.852 $\pm$ 0.049 &  \textbf{0.862 $\pm$ 0.052} &  \textbf{0.809 $\pm$ 0.034} \\
\bottomrule
\end{tabular}
\caption{Comparative study assessing AUC performance of IDC versus ILC subtyping using different self-supervised pretraining methods with varying percentage folds of training data in a 10-fold CV.}
\end{center}
\end{table*}

\begin{table*}
\footnotesize
\begin{center}
\begin{tabular}{ll|rrrrrrrrrr}
\toprule
{} & {} & \multicolumn{8}{c}{\underline{Colorectal Tissue Phenotyping}} \\
Method & SN & Adi &  Back &    Deb &    Lym &    Muc &  Norm &    Str &    Tum &    All \\
\midrule
IN Transfer & Y &  0.983 &  \textbf{1.000} &  0.997 &  0.974 &  0.963 &  \textbf{0.988} &  0.982 &  \textbf{0.978} &  0.983 \\
SimCLR & Y   & 0.988 &  \textbf{1.000} &  0.994 &  0.980 &  0.969 &  0.973 &  0.979 &  0.969 &  0.981 \\
DINO & Y  &  \textbf{0.999} &  \textbf{1.000} &  \textbf{0.999} &  \textbf{0.985} &  \textbf{0.992} &  0.960 &  \textbf{0.992} &  0.967 &  \textbf{0.987} \\
\midrule
IN Transfer & N &  0.988 &  \textbf{0.909} &  0.900 &  0.870 &  0.886 &  \textbf{0.988} &  0.963 &  \textbf{0.978} &  0.935 \\
SimCLR & N  &  0.981 &  0.765 &  0.955 &  \textbf{0.951} &  0.926 &  0.976 &  0.979 &  0.973 &  0.938 \\
DINO & N  & \textbf{0.991} &  0.729 &  \textbf{0.961} &  0.950 &  \textbf{0.978} &  0.957 &  \textbf{0.990} &  0.973 &  \textbf{0.941} \\
\bottomrule
\end{tabular}
\caption{Comparative study assessing AUC performance of self-supervised pretraining on CRC-100K with Macenko stain normalization (SN) (\textbf{top}) and without SN (\textbf{bottom}).}
\vspace{-5mm}
\end{center}
\end{table*}

For all patch-level tasks, we trained K-Nearest Neighbors (KNN) on the extracted embeddings of each pretrained model on each dataset. Though SimCLR and DINO were not pretrained using colorectal tissue, CRC-100K was still included in our study design due to exhibiting distinct morphological phenotypes.

\subsection{Results}

\textbf{ImageNet features are a strong baseline.} Across different tasks, when comparing self-supervised methods against our ResNet-50-B3$_{\text{IN}}$ baseline, we find that ImageNet features achieve slightly lower (but comparable) performance on many tasks. In weakly-supervised learning, ResNet-50-B3$_{\text{IN}}$ achieves an AUC performance of 0.884 on BRCA subtyping, in comparison to the SIMCLR and DINO which achieved 0.879 and 0.886 AUC respectively (\textbf{Table 1}). On patch-level datasets with distinct morphological phenotypes such as CRC-100K (with Macenko normalization), ResNet-50-B3$_{\text{IN}}$ achieves a high AUC of 0.983 (\textbf{Table 2}). One hypothesis for the surprisingly robust $\text{ResNet-50}_{\text{B3, IN}}$ performance is that feature maps before the last residual block are low-level feature descriptors, and thus able to distinguish between clearly distinct morphologies such as tumor versus stroma, or tumor versus adipose tissue. Overall, we find that the gap between self-supervised learning and ImageNet transfer learning is smaller than previously found in other studies. However, performance benefit of self-supervised learning may vary depending on organ type (only breast tissue tested) and weakly-supervised architecture (only CLAM tested). 

\textbf{Self-supervised features are robust and sample-efficient.}
In training weakly-supervised models on less data, we find that self-supervised methods are more robust, as demonstrated in DINO achieving 0.809 AUC with only 25\% of the original training data in BRCA subtyping. In comparison, AUC performance using ResNet-50-B3$_{\text{IN}}$ degrades to 0.756 (\textbf{Table 1}). In patch-level evaluation on more difficult tasks, the performance gap between ResNet-50-B3$_{\text{IN}}$ and self-supervised methods is much wider. On CRC-100K without Macenko stain normalization, the macro-averaged AUC for ResNet-50-B3$_{\text{IN}}$ decreases to 0.935 in comparison to 0.941 from DINO (\textbf{Table 2}). In visualizing UMAP scatter plots of pre-extracted $\text{ResNet-50}_{\text{B3, IN}}$ features, despite the high AUC performance on both CRC-100K datasets, the representation quality is poor as global structures within the same class types are not preserved (\textbf{Figure 4}). On the other hand, global structures for classes such as stroma, tumor, normal, and mucous tissue are well-preserved for self-supervised models. On BreastPathQ, DINO outperforms $\text{ResNet-50}_{\text{B3, IN}}$ on MSE and Kendall-Tau concordance scores with larger performance gaps of 0.029 versus 0.058 MSE and 0.854 versus 0.824 concordance respectively (\textbf{Table 3}).

\textbf{DINO improves over SimCLR.} In comparison with SimCLR used in our ablation study, we observe that DINO also outperforms SimCLR in weakly-supervised tasks (with higher AUC with less finetuning) as well as all patch-level tasks (except lymphocyte classification in CRC-100K without Macenko normalization). In training SimCLR and DINO on the same dataset for pretraining (with default training recipes) and evaluation at 100 epochs, our results suggest that self-supervised learning paradigms such as student-teacher knowledge distillation outperform contrastive-based methods, as constructing positive-negative pairs are not needed in steering representation learning in pathology data.

\begin{figure*}
   \begin{center}
   \includegraphics[width=14cm]{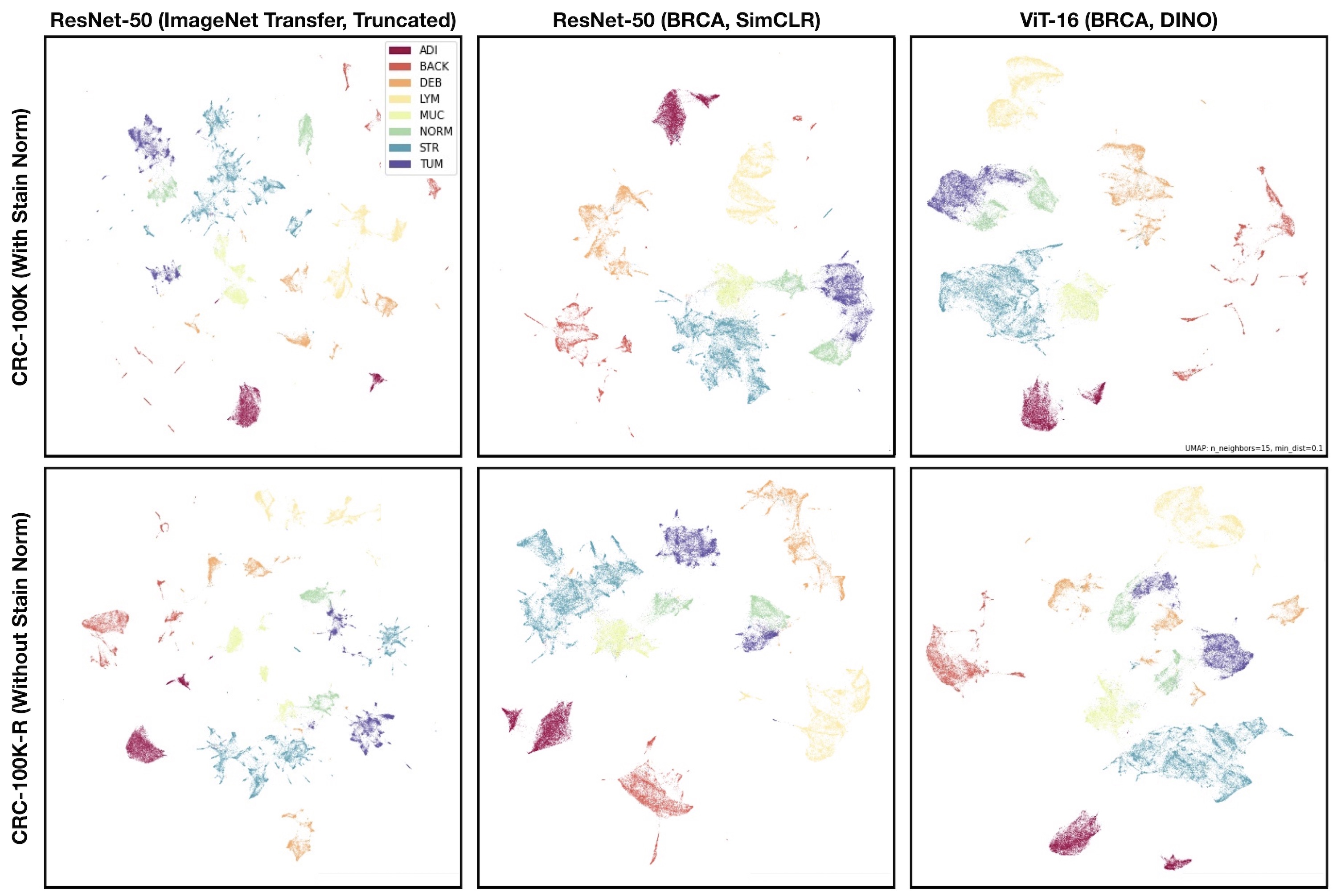}
   \end{center}
   \caption{2D UMAP scatter plot visualizing global structure of extracted feature embeddings of ResNet-50-B3$_{\text{IN}}$, SimCLR, and DINO on CRC-100K. For stroma, tumor, and background tissue classes, DINO is able to preserve the  global structure of these phenotypes in comparison to ResNet-50-B3$_{\text{IN}}$. We used default UMAP parameters of: $\operatorname{neighbors}=15, \operatorname{dist}=0.1$.}
\end{figure*}

\begin{wraptable}{l}{0.5\textwidth}
\begin{tabular}{l | c c }
\toprule
{} & \multicolumn{2}{c}{\underline{Tumor Cellularity Prediction}} \\
Method & MSE $\downarrow$ & Kendall-Tau $\uparrow$ \\
\midrule
IN Transfer &  0.058 & 0.828 \\
SimCLR  &  0.078 & 0.788 \\
DINO &  \textbf{0.029} & \textbf{0.854} \\
\bottomrule
\end{tabular}
\caption{Comparative study assessing MSE and Kendall-Tau concordance score on BreastPathQ.}
\end{wraptable}

\textbf{DINO learns morphological visual concepts.} To introspect into what DINO has learned, we visualize the different attention heads in multi-head self-attention which characterize a normalized distribution over the patch embeddings. Specifically, for a small ViT model in DINO with $n=6$ heads computed on a 256-length sequence of 384-dim patch embeddings, each attention head computes self-attention on different 64-dim segments of the patch embeddings in the sequence. For each head, we use the attention weights from the [CLS] token that pools over the patch embeddings (which focus on different 64-dim segments), which we overlay on the image patch. In visualizing the overlayed attention weights at the last self-attention block (with weights thresholded at 0.5), we note that visualizations of attention distributions each capture distinct morphological phenotypes, localizing cell location, stroma tissue, and fat / air pockets (\textbf{Figure 2, 4}). This observation is in line with current studies that have introspected self-supervised ViT models, in which attention heads can be used as a method for object localization or object discovery in the part-whole hierarchy~\cite{caron2021emerging, simeoni2021localizing}. In the application of histopathology, this introspection reveals that the embeddings learned for the $16 \times 16$ patches capture fine-grained morphological features such as cells and background tissue; we believe this can have applications in the development of automated, interpretable biomarkers.


\section{Conclusion}
As the field migrates towards difficult phenotyping tasks such as biomarker discovery, more domain-specific encoders are needed in elucidating novel morphological biomarkers not currently identified by human pathologists. In this work, we present a comprehensive empirical evaluation of self-supervised pre-training across a diverse variety of weakly-supervised and patch-level tasks. Future work would further assess the utility of self-supervised methods for learning representations of histopathological images from rare diseases for which there is a natural data paucity of slide-level and patch-level annotations.

\section{Acknowledgements}
We thank the BioML group at Microsoft Research New England for their insightful feedback. R.J.C. and R.G.K performed part of this work while at Microsoft Research New England. R.J.C. was supported by the NSF Graduate Fellowship. R.G.K. gratefully acknowledges funding from CIFAR.

\medskip
\small
\bibliographystyle{unsrt}
\bibliography{neurips2021}

\end{document}